\documentclass[runningheads]{llncs}

\usepackage{graphicx}
\usepackage{hyperref}
\usepackage{amsmath,amsfonts}
\usepackage{mathtools}
\DeclarePairedDelimiter\ceil{\lceil}{\rceil}
\usepackage{algorithm}%
\usepackage{algorithmicx}%
\usepackage{algpseudocode}%
\usepackage{subfig}
\usepackage[labelfont=bf]{caption}
\usepackage{xcolor}
\usepackage{array}
\usepackage{multirow}
\usepackage{caption}
\begin{document}
\title{CoProNN: Concept-based Prototypical Nearest Neighbors for Explaining Vision Models}
\titlerunning{CoProNN}
\author{Teodor Chiaburu\inst{1}\orcidID{0009-0009-5336-2455} \and
Frank Haußer\inst{1}\orcidID{0000-0002-8060-8897} \and
Felix Bießmann\inst{1,2}\orcidID{0000-0002-3422-1026}}
\authorrunning{T. Chiaburu et al.}
\institute{Berliner Hochschule für Technik, Berlin, Germany\\
\email{\{chiaburu.teodor;frank.hausser;felix.biessmann\}@bht-berlin.de} \and
Einstein Center Digital Future, Berlin, Germany}
\maketitle              % typeset the header of the contribution
\begin{abstract}
Mounting evidence in explainability for artificial intelligence (XAI) research suggests that good explanations should be tailored to individual tasks and should relate to concepts relevant to the task. However, building task specific explanations is time consuming and requires domain expertise which can be difficult to integrate into generic XAI methods. A promising approach towards designing useful task specific explanations with domain experts is based on compositionality of semantic concepts. Here, we present a novel approach that enables domain experts to quickly create concept-based explanations for computer vision tasks intuitively via natural language. Leveraging recent progress in deep generative methods we propose to generate visual concept-based prototypes via text-to-image methods. These prototypes are then used to explain predictions of computer vision models via a simple k-Nearest-Neighbors routine. The modular design of CoProNN is simple to implement, it is straightforward to adapt to novel tasks and allows for replacing the classification and text-to-image models as more powerful models are released. The approach can be evaluated offline against the ground-truth of predefined prototypes that can be easily communicated also to domain experts as they are based on visual concepts. We show that our strategy competes very well with other concept-based XAI approaches on coarse grained image classification tasks and may even outperform those methods on more demanding fine grained tasks. We demonstrate the effectiveness of our method for human-machine collaboration settings in qualitative and quantitative user studies. All code and experimental data can be found in our GitHub repository: \href{https://github.com/TeodorChiaburu/beexplainable}{https://github.com/TeodorChiaburu/beexplainable}.

\end{abstract}

%%% INTRODUCTION %%%
\section{Introduction}\label{sec1}

Empirical evidence in the field of eXplainable Artificial Intelligence (XAI) suggests that good explanations for Machine Learning (ML) predictions should be designed for each task individually \cite{poursabzi-sangdehManipulatingMeasuringModel2018,lageEvaluationHumaninterpretabilityExplanation2019,haseEvaluatingExplainableAI2020a,Kim2022HIVE}. This requires understanding of both the ML model and the domain of the respective task. These findings highlight the importance of human experts in the development of XAI methods.
Integrating knowledge of domain experts in XAI methods is challenging. Not only because of potential cognitive bias \cite{herman2017}, but also because of the time needed to design task specific XAI methods. Especially when dedicated training and expert knowledge is needed for a task, such as in classification problems with large label spaces, rare classes and small differences between classes, it is often difficult to scale the human part in the XAI algorithm development.

To address these challenges, we propose a novel approach to XAI by leveraging recent advancements in multimodal foundation models and deep generative models. In order to enable human experts to easily develop explanations in natural language, we make use of text-to-image methods, in particular Stable Diffusion \cite{stabdiff}.  Importantly, the proposed approach allows to improve the efficiency of human experts when developing explanations for a novel task and when evaluating individual XAI methods. We demonstrate  the effectiveness of our approach in experiments on heterogeneous tasks requiring domain expertise for designing explanations. Our results show that CoProNN is simple to adapt to a broad range of tasks. Compared to existing similar XAI approaches our methodology compares favourably both in terms of offline metrics as well as in qualitative and quantitative user studies. 

We note that we aimed at embodying into our method the benefits of two worlds: concept- and prototype-based explanations. As discussed in the following sections, the common approach for concept-related methods is to infer semantic significance from a pool of many general pre-annotated concepts, while prototypical models search for representative fragments/patches in the training samples that are meant to capture the defining traits of the class prototypes. To the best of our knowledge, no other XAI framework is currently designing task-specific prototypes separate from the training set, based on higher level concepts.

%%% RELATED WORK %%%
\section{Related Work}\label{sec:related_work}

Early attempts at post-hoc XAI approaches focused on feature attribution maps that identify relevant pixels in the input image \cite{salmaps1,salmaps2,salmaps3,salmaps4,salmaps5} or visualizing what features the network learns during training \cite{fviz}. These methods have the key advantage that they do not require task specific adaptations. As such, these methods are simple to scale to a large number of tasks with minimal manual effort. On the other hand, they can be difficult to use in practice, as they rarely provide useful explanations other than some localization of features relevant for a given prediction. Another challenge with most attribution based post-hoc explanation methods is that attribution maps are often noisy, non-robust and unfaithful to the predictive behavior of the model \cite{salcritique1,salcritique2,salcritique3,salcritique4}. 

Complementing this work, other researchers have focused on concept-based explanations, leveraging principles of semantic compositionality of intuitive visual \textit{concepts} \cite{tcav,ace,Chen_2020,koh2020concept,fel2023craft,Zhou_2018_ECCV,ramaswamy2022elude}, \textit{prototypes} \cite{protopnet,donnelly2022deformable,nauta2021neural,rymarczyk2022interpretable}, \textit{examples from the training set} \cite{koh2020understanding,yeh2018representer,NEURIPS2021_c460dc0f} or \textit{captioning} \cite{hendricks2016generating,sammani2022nlxgpt}, in order to generate  explanations that are comprehensible for human subjects. Here, we build on this work in order to combine the scalability of post-hoc attribution methods with the intuitive comprehensibility of semantic explanations by leveraging recent progress in generative deep learning methods. 
Our contribution is similar to prior work on \textit{prototype methods} and \textit{concept-based methods}. Prototype methods usually search for prototypes or prototypical parts in the training set of the classifier and often aim at upgrading the given classifier into an interpretable model by retraining it. This retraining step is usually not necessary for post-hoc explanation methods, like CoProNN. Another disadvantage of classical prototype methods is that integrating task relevant concepts is often not straightforward. Such concepts, however, can be useful if not required to relate an explanation to expert knowledge for a given task. 
Two popular post-hoc methods to integrate domain expertise via task relevant concepts into explanations are \textbf{T}esting with \textbf{C}oncept \textbf{A}ctivation \textbf{V}ectors (TCAV) \cite{tcav} and \textbf{I}nterpretable \textbf{B}asis \textbf{D}ecomposition (IBD) \cite{Zhou_2018_ECCV}.

TCAV fits a linear classifier in one of the model's layers, that separates a concept's examples from random counterexamples. Hence, the CAVs are the vectors orthogonal to these decision boundaries. To quantify how sensitive the class prediction is towards each concept, TCAV computes the directional derivatives of the class members w.r.t. the CAVs multiple times for different partitions of the random dataset. The TCAV score for a whole class w.r.t. a concept is defined as the fraction of class members having positive directional derivatives throughout all of that concept's CAVs. For a single sample, the score would only count the positive derivatives throughout the concept's CAVs (not class-wise).

IBD decomposes the neural activations of the input into semantically interpretable components. It achieves this by fitting logistic regressors to separate the concept examples from counterexamples. The resulting 'concept weight vectors' will construct a concept basis onto which the rows in the Dense layer matrix responsible for predicting the class probabilities (called 'class weight vectors') are projected. The resulting class concept scores represent the concept contributions to the prediction of a classifier. The IBD concept relevance scores for a single sample are the dot products between the sample's feature vector and each of the concept components that approximate its class weight vector, see Appendix~\ref{ap:explainer}.

%%% METHODS %%%
\section{Methods}

In this section we first describe the overall workflow of the proposed approach followed by a more detailed description of the key components and conclude with presenting the datasets and evaluation methods. 

\subsection{CoProNN}

\begin{figure*}[h!]
    \begin{center}
       \includegraphics[width=1.0\textwidth]{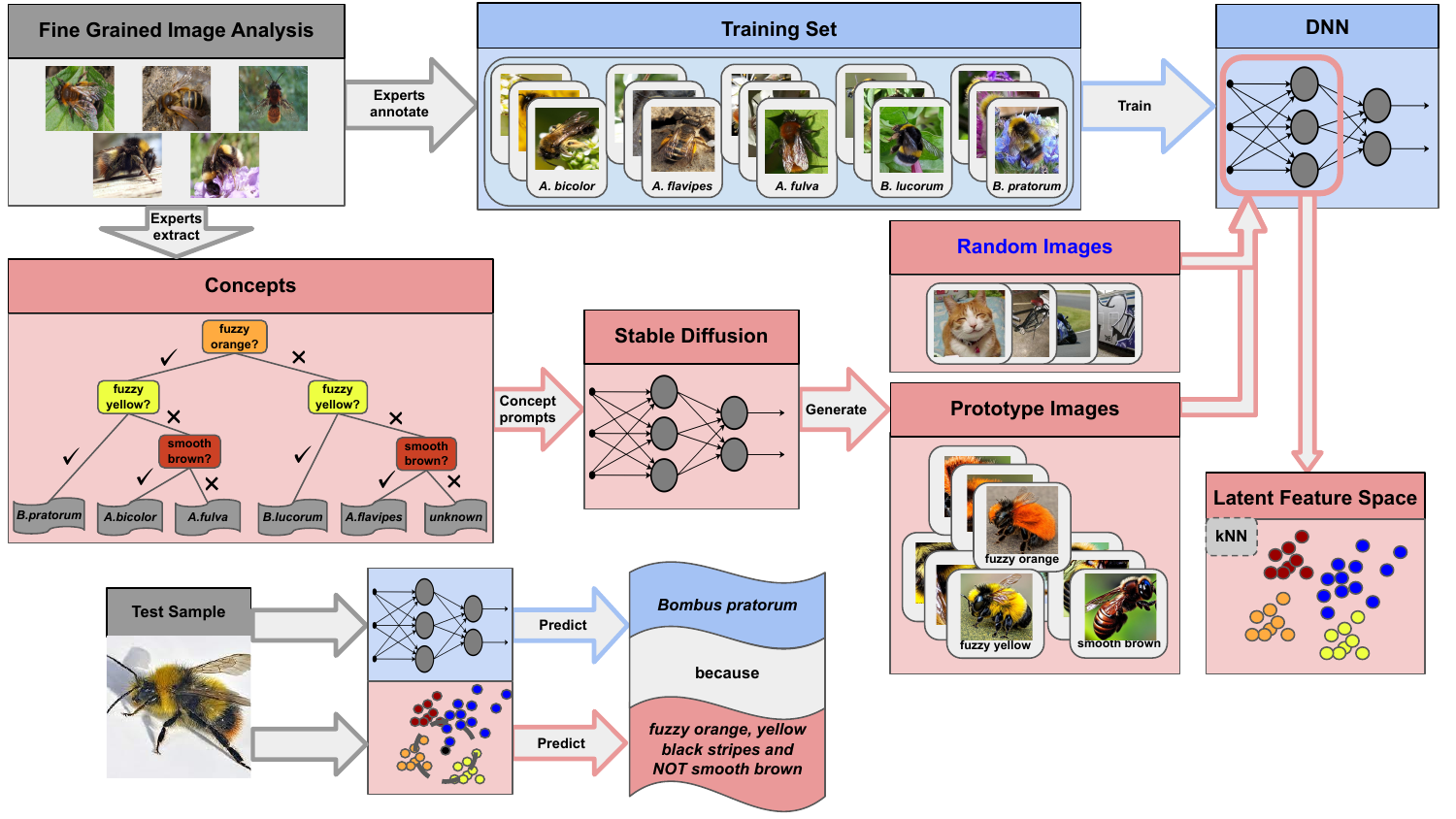}
    \end{center}
    \caption{Overview of the CoProNN framework, illustrated on the subset of wild bee images scraped from \textit{iNaturalist}.  \textbf{Classifier} (\textit{blue boxes}): images of the five pre-selected species are fed along with their labels assigned by entomologists into a standard DNN. \textbf{Concept based explanations} (\textit{red boxes}): domain experts define a set of intuitive concepts to discriminate the five species in a hierarchical manner (the decision tree block). These concepts are used as prompts to generate prototype images with Stable Diffusion. The prototype images along with a set of random images are mapped into the latent feature space of the DNN's frozen backbone, where they will be fed into kNN. At inference (\textit{bottom row}): the DNN classifier predicts a species label and an explanation is retrieved based on the kNN-computed similarities between the extracted features of the test sample and the prototype images in the latent space.}
\label{fig:pipeline}
\end{figure*}

Figure \ref{fig:pipeline} depicts the inner workings of the CoProNN approach, which consists of two main stages:

\begin{enumerate}
    \item[1)] \textbf{Training the Classifier}: A Deep Neural Network (DNN) is trained (or a standard computer vision backbone is fine tuned) on the given dataset of images until convergence. For our experiments, we have used a ResNet50 \cite{resnet}, but any DNN able to extract image features can be inserted. More details about the training process can be found in the Appendix~\ref{ap:explainer}.
    \item[2)] \textbf{Fitting the Post-Hoc Explanation Method}: Depending on the knowledge and skills required to solve the classification task, domain experts define a minimal set of relevant concepts a layman would need to be able to classify the images. For example, for our wild bee dataset, entomologists came up with the simple taxonomic tree sketched in Figure \ref{fig:pipeline}. The nodes represent intuitive visual concepts related to the color and the texture of the bees' body parts. In a hypothetical medical application, radiologists could similarly draw a decision tree to classify different types of cancerous cells, using features such as cell shape or size. Once the concepts are agreed upon, they are turned into prompts fed into a generative text-to-image model that produces prototypical images for each concept. These prototypes are passed through the frozen backbone of the previously trained DNN and mapped into its latent feature space. A set of random images is similarly transformed and fed along with the prototypical vectors into a kNN. The check against random images ensures that CoProNN is also able to predict the absence of all the concepts in an image.
\end{enumerate}

The explanations are generated post-hoc by comparing the test sample in the feature space of the classifier against the prototype and the random vectors. The kNN probabilities are interpreted as relevance scores for the prototypes. Concepts associated with a prediction are presented to the end-user next to the DNN label prediction. The general explanation format is, hence: \textit{"This image is class A, because concepts X, Y ... are present and concepts Z, W ... are absent"}. To expand the explanation modality, the textual formulation can be accompanied by one prototype image per concept. More details on the generation of the prototypes and the prediction of the explanation follow in the next subsections.

\subsection{Prototype Images via Stable Diffusion}
\label{ss:SD}
For creating the prototypical images characterized by representative concepts, a Stable Diffusion (SD) v1.5 model from the \textit{diffusers} library \cite{diffusers} in PyTorch was used. A key advantage of modern generative models is that these concepts can be retrieved by human domain experts using short natural language prompts for the diffusion model. For instance, a well known animal such as \textit{cheetah} is simple to describe with intuitive concepts i.e. 'animal with yellow fur and black dots'. 

When transitioning to a more fine-grained context, expert knowledge will often be required. For example, the basic concepts to distinguish the five wild bees (Fig. \ref{fig:cognition}) were proposed by the entomologist we collaborated with. By following his suggestions and the prompting guidelines in \cite{prompts1,prompts2,prompts3}, we came up with the successful prompts for generating the prototypes. Their shorthand version can be found in Fig. \ref{fig:ex_protos}. More details on the extra prompt modifiers that we used to increase image resolution and fidelity can be found in our code repository.

\subsection{Nearest Neighbors as Explanations}

We leverage k Nearest Neighbors (kNN) in the feature space of a DNN classifier to retrieve the concepts relevant to class prediction. Other XAI methods \cite{nguyen2023visual,Keane_2019,papernot2018deep} also make use of kNN to explain a model's decision. However, in contrast to CoProNN, their search space for explanations resides in the training set of images the classifier was trained on. Algorithm \ref{alg:knn_concepts} in the Supplement describes our approach to formulating explanations. 

For each concept $j$, $1 \leq j \leq m$, all prototype images obtained via SD as described in Section~\ref{ss:SD} along with  a set of random images 
are mapped into the feature space of the last layer of the  trained DNN classifier. We denote by $\Omega_j$ the set of feature vectors of the prototype images for concept $j$.

Given a new image $x$, we first compute a feature vector $f(x)$  (the activation of the last layer of the DNN for image $x$) and pass this feature vector to the kNN model trained on the concept prototype images to obtain the likelihood of that feature vector $f(x)$ being associated with each concept image set $\Omega_j$.

In order to improve the robustness of the kNN model we also include a set of random images of the same size as each of the prototype image sets.
We randomly sample the set of random images as a subset/partition of a large image set, kNN is run for all the different random partitions and the scores $p_j$ are averaged. While this inclusion of a random concept image set would not be required, we perform this step for the sake of a fair comparison with the TCAV and IBD studies. Both of these methods we compare with incorporate a set of counterexamples paired with positive concept examples.

To decide which of the $m$ concepts are shown to the end-user as relevant, either the top-N predicted concepts (sorted according to their kNN scores $p$) are selected, or a threshold $t \in (0, 1]$ is defined; only prototype sets for which $p_{j} \geq t$ would then be considered for formulating an explanation of the type \textit{"the model predicts class A because concept X is present and concept Y is absent."} Note that the absence of a concept is also of importance for hierarchical classification routines like the decision tree drawn in Figure \ref{fig:pipeline}. The decision for top-N or threshold based concept predictions is similar to that of multiclass vs multilabel paradigms. CoProNN can be readily adapted to both settings, depending on the application specific requirements, such as the question of how many concepts should be used for an explanation.

\subsection{Coarse and Fine Grained Classification Tasks}\label{sec:datasets}

To illustrate the effectiveness of CoProNN to generate task-specific explanations via concepts relevant to a given domain we employ three different data sets:

\begin{itemize}
    \item[1)] \textbf{Animals from ImageNet} \cite{imagenet_cvpr09}: only a subset of easily distinguishable classes was downloaded: \textit{cheetah, garter snake, tiger, mud turtle, zebra} (150 images each). The choice of these classes was inspired by the experiments carried out in the paper introducing TCAV \cite{tcav}, where the authors tested their algorithm on straightforward concepts such as 'striped', 'dotted' or 'chequered'. These concepts also made the key words in the prompts for generating the concept based prototypes (one prompt per class, see Fig.~\ref{fig:ex_protos} (a) - (e)).
    \item[2)] \textbf{Wild Bees from iNaturalist} \cite{iNat}: 30k images of the top 25 most frequent wild bee species with their natural habitat in Germany were scraped from the iNat online database\footnote{\href{https://www.inaturalist.org/}{https://www.inaturalist.org/}}. To keep label noise in check, only images awarded a 'research grade' (and possessing a CC-BY-NC copyright license) were downloaded. The 25 species were identified in collaboration with our entomologist. After preliminary experiments, we have reduced the subset down to five particularly difficult and frequently confounded species: \textit{Andrena bicolor/flavipes/fulva} and \textit{Bombus lucorum/pratorum} (see Fig. \ref{fig:cognition}). This resulted in a dataset totalling 7595 images of wild bees. The concept keywords used to distinguish the five bee species (also highlighted in the decision tree in Fig. \ref{fig:pipeline}) are: 
    \begin{enumerate}
        \item[*] 'fuzzy orange' - \textit{A. fulva} (completely orange), \textit{A. bicolor} (orange thorax) and \textit{B. pratorum} (orange sting region)
        \item[*] 'fuzzy yellow with black stripes' - both \textit{Bombus} on thorax and abdomen
        \item[*] 'smooth shiny dark brown' - \textit{A. bicolor} and \textit{A. flavipes} on abdomen.
    \end{enumerate}
    The prompts for generating the corresponding concept based prototypes and some examples are given in Fig.~\ref{fig:ex_protos} (f) - (h). 
    \item[3)] \textbf{Broden} \cite{bau2017network}: the standard dataset of pre-annotated concept images; used for comparison against our approach defining task-specific concepts and for sampling the set of random images (1000 samples).
\end{itemize}

\def\wExProtos{0.18}
\begin{figure*}[h!]
    \centering
    \subfloat[animal with yellow fur and black dots]{\includegraphics[width=\wExProtos\textwidth]{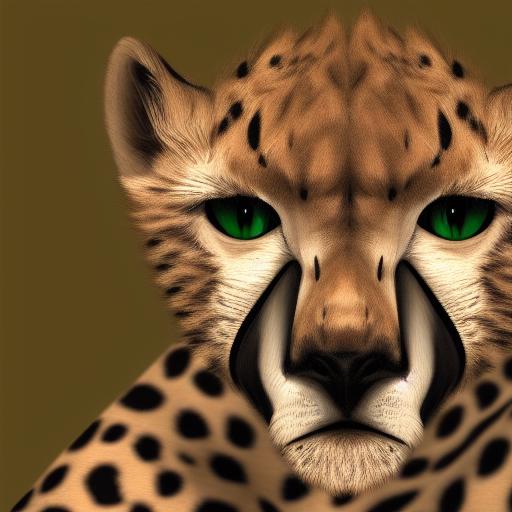}}\hfill
    \subfloat[elongated chequered reptile]{\includegraphics[width=\wExProtos\textwidth]{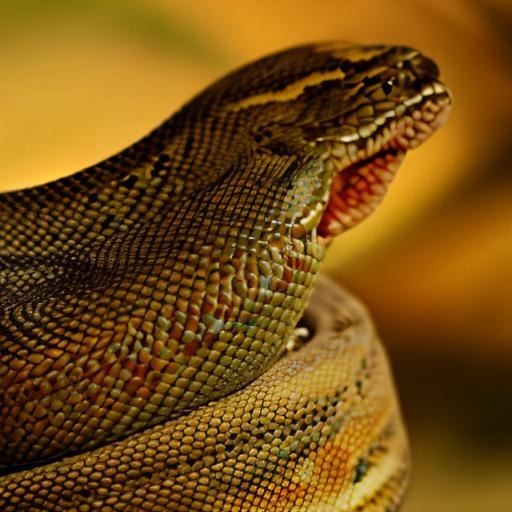}}\hfill
    \subfloat[animal with orange fur and black stripes]{\includegraphics[width=\wExProtos\textwidth]{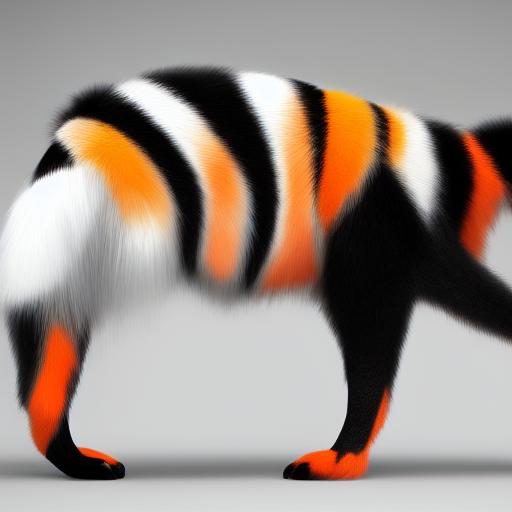}}
    \hfill
    \subfloat[reptile with chequered shell]{\includegraphics[width=\wExProtos\textwidth]{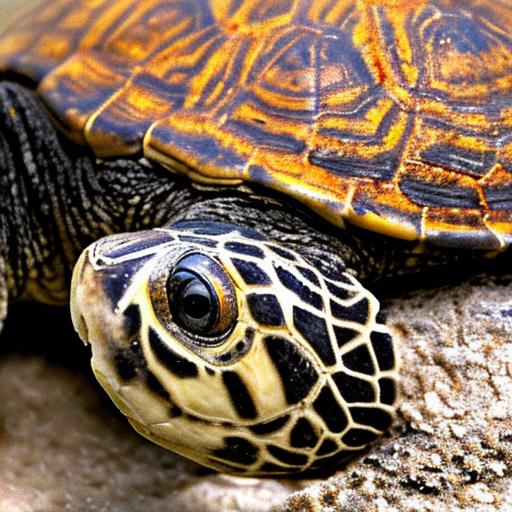}}\hfill
    \subfloat[white animal with black stripes]{\includegraphics[width=\wExProtos\textwidth]{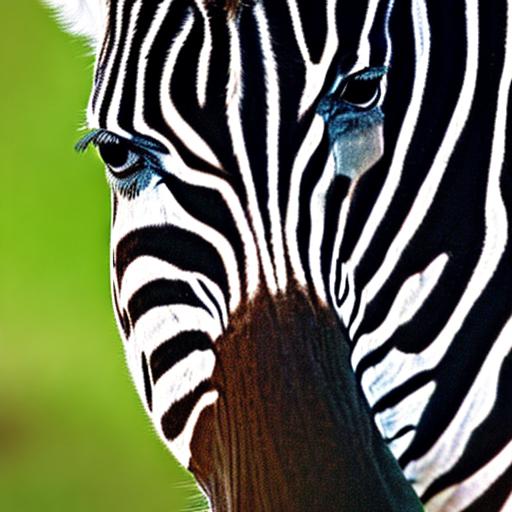}}
    
    \subfloat[bee with fuzzy yellow and black stripes]{\includegraphics[width=\wExProtos\textwidth]{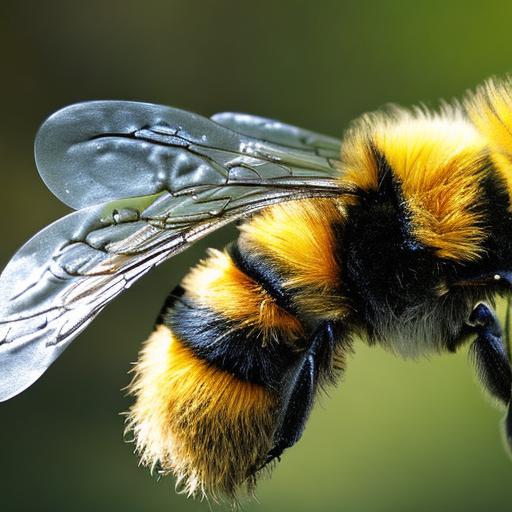}}\hfill
    \subfloat[smooth shiny dark brown bee]{\includegraphics[width=\wExProtos\textwidth]{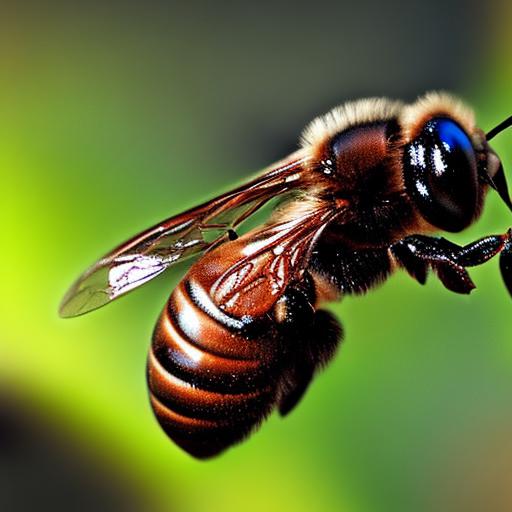}}\hfill
    \subfloat[fuzzy dark orange bee]{\includegraphics[width=\wExProtos\textwidth]{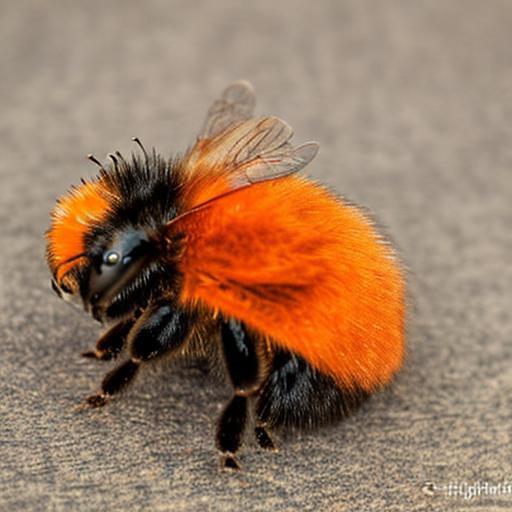}}
    \caption{Examples of prototype images generated by SD along with their (abbreviated) prompts underneath. Prototypes (a) to (e) correspond to the ImageNet animal classes, prototypes (f) to (h) to the iNat wild bees.}
    \label{fig:ex_protos}
\end{figure*}

Note the distortions and unrealistic appearance of the examples in Figure \ref{fig:ex_protos}. To fix this, one could simply generate concept images by using very straightforward prompts such as "zebra, highly detailed". The generated images would resemble very much real images of the considered class, improving the kNN search. However, this would miss the point of our work. The aim here is to show that using concept key words such as "white animal with black stripes" and generating not perfectly looking prototypes allows our method to generalize better without sacrificing accuracy (more details in Section \ref{sec:results}). In the case of finer grained classes, such as the wild bees, a text-to-image model would struggle to understand a prompt devoid of context such as "Andrena fulva" (this would just generate a common honey bee).

\subsection{Evaluation without Humans-in-the-Loop}

All XAI methods considered in this work output a vector of relevance scores for every concept. One key advantage of concept-based explanations is that they can be tested against ground-truth labels of concepts associated with a category. As an example, we know that zebras are white with black stripes or that \textit{A. bicolor} bees are both fuzzy orange and shiny dark brown. We can, therefore, encode the true explanation labels as a binary vector, so that, for instance, an \textit{A. bicolor} bee is represented as a vector $[1, 0, 1]$ - translated as \textit{IS fuzzy orange, IS NOT fuzzy yellow, IS shiny brown}. 

To investigate to what extent the predicted concept relevance vectors match the ground truth concept-label association, we measure their similarity via Cosine Similarity\footnote{\href{https://scikit-learn.org/stable/modules/metrics.html\#cosine-similarity}{https://scikit-learn.org/stable/modules/metrics.html\#cosine-similarity}} (CS) sample-wise and report the averaged values per class. We are using CS for comparing the methods (and not Euclidean distance, for instance) so that only the direction between the true and predicted explanation vectors is taken into consideration. Magnitude should not play a role, since i.e. scores in the TCAV explanation vectors can each range between 0 and 1, while scores in the CoProNN and IBD explanations range cumulatively (when added up) from 0 to 1 (see Fig. \ref{fig:imagenet_bees_proto}).

\subsection{Quantitative User Study}
We evaluated whether our proposed explanation method facilitates human-machine collaboration in a web based user study using \textit{jspsych}~\cite{jspsych}. The app was deployed on the platform \href{https://www.cognition.run/}{https://www.cognition.run}; a total of 80 subjects took part in our investigation. The participants were informed that the data gathered would solely be used for research purposes. The subjects' identities were anonymous and before starting they were given a detailed description of what they were required to do. The experiment was approved by our institutional research board. 

Subjects were randomly divided into two groups: Users in the \textbf{Control Group} would only receive the model predictions as an aid for solving the tasks, but not the explanations, and users in the \textbf{CoProNN Group} could see the model predictions as well as the CoProNN explanation for the prediction. The task interface for the CoProNN users is shown in Fig. \ref{fig:cognition} in the Supplement; a demo can be inspected at: \href{https://hgyl4wmb2l.cognition.run}{https://hgyl4wmb2l.cognition.run}.
Users are presented with a random selection of 10 out of a pool of 24 image samples from the iNat wild bee dataset. The images were selected such that every user would be shown 2 images wrongly classified by our model (out of 4 total misclassifications in the pool of 24). The remaining 8 images were samples correctly classified by our model. The 4 misclassified samples were also chosen such that the CoProNN-explanation would match the wrong species predicted by the model. 
%For instance, a \textit{B. lucorum} which is fuzzy and has yellow stripes is classified as an \textit{A. flavipes} because it is presumably shiny dark brown. 
Since the wild bees can have more than one correct concept label - in the case of \textit{A. bicolor} and \textit{B. pratorum} - a threshold $t = 0.4$ was chosen to decide which concepts to show the users (refer to Algorithm \ref{alg:knn_concepts}). With $k = 10$ neighbors to choose from\footnote{Value set before hypertuning, more details in the Appendix.} and $m = 3$ concepts competing against one another, the number $0.4 = \frac{1}{k}\ceil{\frac{k}{m}}$ appeared to be a good default threshold.

Based on a pilot study, we determined a lower bound of 4 images, corresponding to the 20\% percentile in the distribution of correct answers, that should be classified correctly by each user. Subjects who did not meet this requirement did not qualify for analysis of the user study. In total, this resulted in 41 subjects in the Control Group and 34 in the CoProNN Group.

\subsection{Qualitative User Study}
To better understand qualitatively how subjects used and perceived the explanations provided, we asked users in the CoProNN Group to fill in a standard survey, the \textit{System Causability Scale}  \cite{holzingerMeasuringQualityExplanations2020}. Subjects in the CoProNN Group were asked to answer the following questionnaire items with one of five responses that range from 'strongly agree' to 'strongly disagree':

\begin{enumerate}
\item \textit{I did not need support to understand the model's explanations.}
\item \textit{I found the model's explanations helped me to understand causality.}
\item \textit{I was able to use the model's explanations with my knowledge base.}
\item \textit{I think that most people would learn to understand the model's explanations very quickly.}
\end{enumerate}

%%% RESULTS %%%
\section{Results}\label{sec:results}
In this section we describe the results from our quantitative and qualitative comparisons to existing concept-based XAI methods (TCAV and IBD). More details on how exactly TCAV and IBD were applied to our use-cases can be found in the Supplement~\ref{ap:explainer}.

\subsection{Explanations via Task-Specific Concept-Based Prototypes}

We formulated higher-level concept-based prototype images in SD, tailored specifically for the tasks at hand. Figure \ref{fig:ex_protos} shows examples of such prototypes for every class in the two datasets along with the (abbreviated) prompts used to generate them in SD.

\def\wPlotsProtos{0.33}
\begin{figure*}[h!]
    \centering
    \subfloat[CoProNN]{\includegraphics[width=\wPlotsProtos\textwidth]{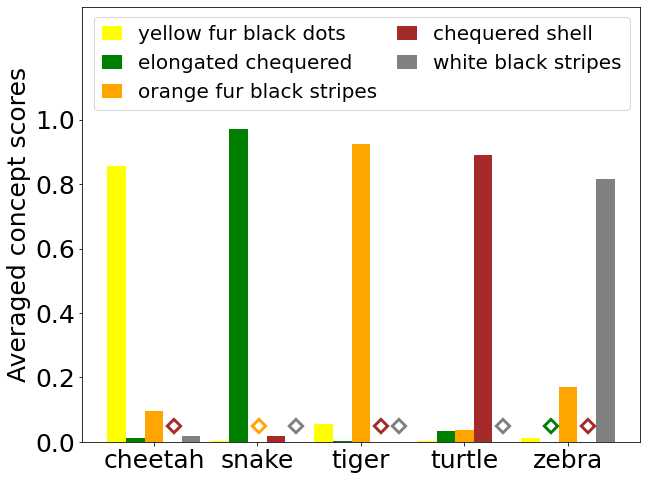}}
    \subfloat[TCAV]{\includegraphics[width=\wPlotsProtos\textwidth]{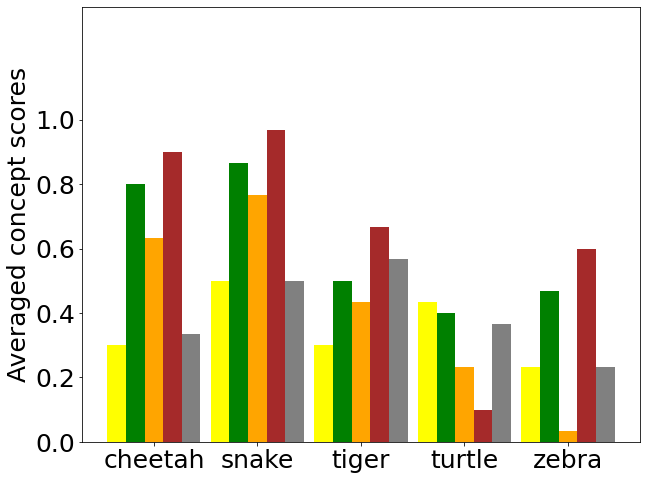}}
    \subfloat[IBD]{\includegraphics[width=\wPlotsProtos\textwidth]{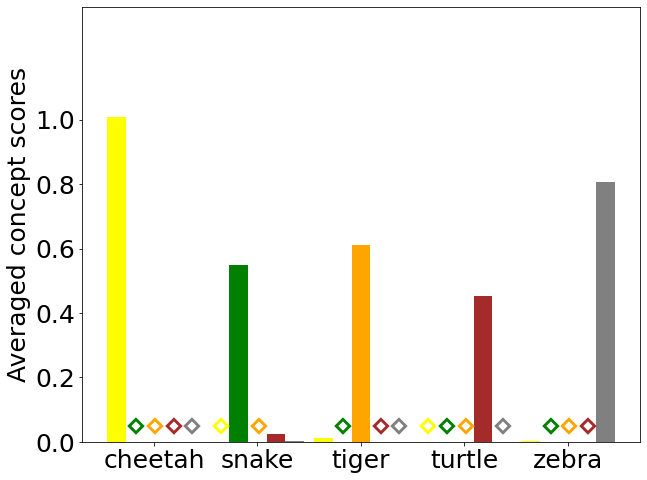}}
    \hfill
    \subfloat[CoProNN]{\includegraphics[width=\wPlotsProtos\textwidth]{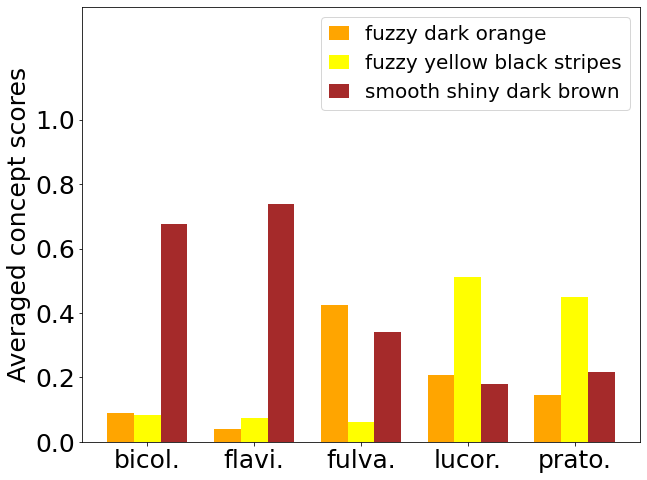}}
    \subfloat[TCAV]{\includegraphics[width=\wPlotsProtos\textwidth]{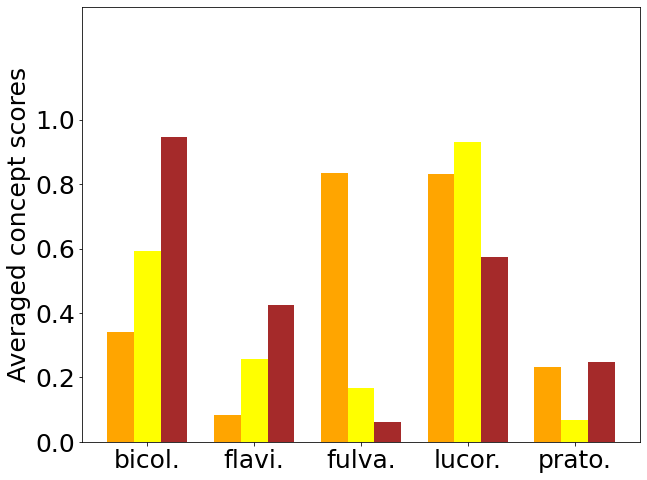}}
    \subfloat[IBD]{\includegraphics[width=\wPlotsProtos\textwidth]{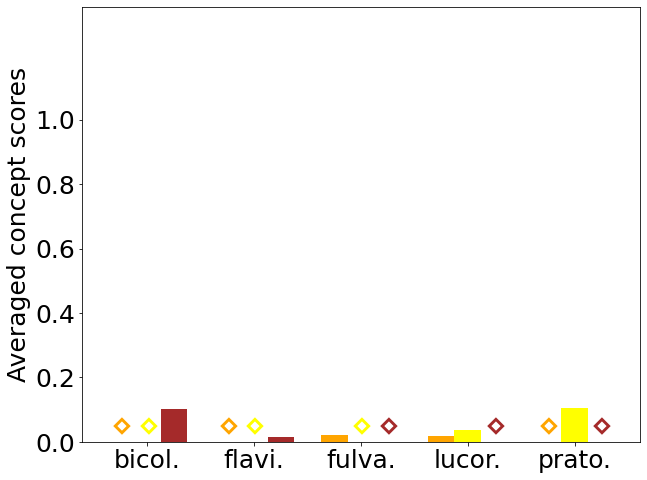}}
    \hfill
    \caption{CoProNN finds the relevant concepts for discriminating categories better than alternative XAI methods (TCAV and IBD). Top row (plots (a) to (c)) depicts the five ImageNet animal classes. The expected explanations are: "yellow fur black dots" for \textit{cheetah}, "elongated chequered" for \textit{snake}, "orange fur black stripes" for \textit{tiger}, "chequered shell" for \textit{turtle} and "white black stripes" for \textit{zebra}. Bottom row (plots (d) to (f)) refers to the iNat wild bees. The expected explanations are (in shorthand): "orange" and "brown" for \textit{A. bicolor}, "brown" for \textit{A. flavipes}, "orange" for \textit{A. fulva}, "yellow" for \textit{B. lucorum}, "yellow" and "orange" for \textit{B. pratorum}. A diamond $\diamond$ marks a zero-value. All the plots display the concept relevance scores averaged per class. An extra note on the normalization of the IBD scores can be found in the Appendix and the corresponding concept contribution scores for the whole classes are given in Fig. \ref{fig:ibd_cls_scores}. Our CoProNN method identifies relevant concepts with high certainty in both datasets, which is made clear by the peaks in plots (a) and (d).}
    \label{fig:imagenet_bees_proto}
\end{figure*}

The results in Figure \ref{fig:imagenet_bees_proto} demonstrate that prototypical images capturing higher-level concepts enable CoProNN to retrieve more faithful and robust explanations\footnote{We have verified the robustness and faithfulness of our explanations by running multiple iterations w.r.t. various random partitions. More advanced techniques such as adversarial attacks could also be employed.} than TCAV and IBD. Ideally, the ground-truth concept(s) for every species should achieve high relevance scores and should consistently stand out as the most relevant of the predicted concepts. Note that IBD allows for negative relevance scores,  meaning that the concept is considered irrelevant for that class; for the sake of comparability we mapped IBD negative scores to 0 when displaying them in Figures \ref{fig:imagenet_bees_proto}, \ref{fig:imagenet_con} and computing the metrics in Tables \ref{table:protos}, \ref{table:concepts}.

The ImageNet classes, indeed, exhibit high relevance scores when explained by CoProNN and IBD (see Fig. \ref{fig:imagenet_bees_proto} a, c), with CoProNN attaining higher certainty in assigning scores to the relevant concepts. On the other hand, TCAV-computed relevance scores do not single out relevant concepts for any of the classes, oftentimes even predicting the wrong concept (see Fig. \ref{fig:imagenet_bees_proto} b). 

Also on the wild bee dataset, CoProNN assigns high relevance scores to the correct concepts (see Fig. \ref{fig:imagenet_bees_proto} d). We notice, however, that species requiring more than one concept to be uniquely characterized remain difficult to fully explain: \textit{A. bicolor}, which is both fuzzy orange and shiny dark brown and \textit{B. pratorum}, which is mainly fuzzy yellow with black stripes, but also has a fuzzy orange terminal segment on the abdomen. For both of them, CoProNN confidently predicts only one of the concepts. For \textit{A. bicolor}, TCAV does assign more importance to the secondary concept 'fuzzy orange' but also to the incorrect concept 'fuzzy yellow'. This culminates in predicting with highest confidence the single wrong available concept for \textit{B. pratorum}, namely 'shiny brown'. IBD has the most difficulty in recognizing relevant concepts for the wild bees.

The CS values in Table \ref{table:protos} stem from comparing the ground truth concepts associated with a class with the predicted concept relevance scores. CoProNN compares very well against TCAV and IBD on the animal classes from the standard (overfitted by many XAI methods) coarse grained classification dataset ImageNet (all scores above 0.9). On more difficult fine grained tasks in the wild bee dataset, our method outperforms the competitors in explaining most classes.

\setlength{\tabcolsep}{10pt} % increase cell padding and stretch rows
\renewcommand{\arraystretch}{1.5}
\begin{table}[h!]
    \centering
    \begin{tabular}{|m{5.5em} || m{8em} | m{8em} | m{8em} |} 
     \hline
      & \multicolumn{3}{|c|}{\textbf{Cosine Similarity}} \\
     \hline
     \textbf{\textit{Class}} & \textbf{CoProNN} & \textbf{TCAV} & \textbf{IBD} \\
     \hline
     \textit{Cheetah} & 0.9177 $\pm$ 0.2126 & 0.2306 $\pm$ 0.001 & \textbf{1.0 $\pm$ 0.0} \\
     \hline
     \textit{Snake} & 0.9956 $\pm$ 0.022 & 0.5101 $\pm$ 0.001 & \textbf{0.9986 $\pm$ 0.0017} \\
     \hline
     \textit{Tiger} & \textbf{0.9715 $\pm$ 0.1509} & 0.3953 $\pm$ 0.001 & 0.9532 $\pm$ 0.3013 \\
     \hline
     \textit{Turtle} & 0.9589 $\pm$ 0.103 & 0.2203 $\pm$ 0.001 & \textbf{1.0 $\pm$ 0.0} \\
     \hline
     \textit{Zebra} & 0.9364 $\pm$ 0.1022 & 0.0709 $\pm$ 0.001 & \textbf{1.0 $\pm$ 0.0} \\
     \hline\hline
     
     \textit{A. bicolor} & \textbf{0.7831 $\pm$ 0.0763} & 0.7142 $\pm$ 0.1527 & 0.66 $\pm$ 0.2539 \\ 
      \hline
     \textit{A. flavipes} & \textbf{0.9926 $\pm$ 0.0043} & 0.667 $\pm$ 0.3386 & 0.6667 $\pm$ 0.7454 \\
     \hline
     \textit{A. fulva} & 0.7373 $\pm$ 0.2118 & 0.9223 $\pm$ 0.1183 & \textbf{0.9247 $\pm$ 0.3596} \\
     \hline
     \textit{B. lucorum} & \textbf{0.7929 $\pm$ 0.2027} & 0.6549 $\pm$ 0.0827 & 0.7644 $\pm$ 0.4016 \\
     \hline
     \textit{B. pratorum} & \textbf{0.7639 $\pm$ 0.1412} & 0.5882 $\pm$ 0.144 & 0.66 $\pm$ 0.2539 \\
     \hline
    \end{tabular}
    \caption{CoProNN reliably finds concepts relevant for a given class. Cosine similarity (CS) is computed between the true concept labels and the predicted concept relevance scores from each method. A high CS outlines a concept relevance vector very similar to the concept ground-truth. CoProNN outperforms competitors TCAV and IBD on 4 out of 5 wild bee classes and achieves very high scores on the ImageNet animal classes (above 0.9, at least second best).}
    \label{table:protos}
\end{table}

\subsection{Explanations via Task-Unspecific Concepts}

\begin{figure*}[ht!]
    \centering
    \subfloat[CoProNN]{\includegraphics[width=\wPlotsProtos\textwidth]{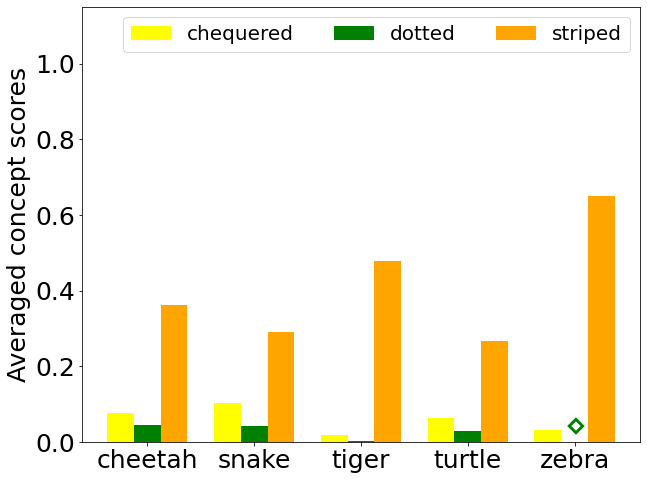}}
    \subfloat[TCAV]{\includegraphics[width=\wPlotsProtos\textwidth]{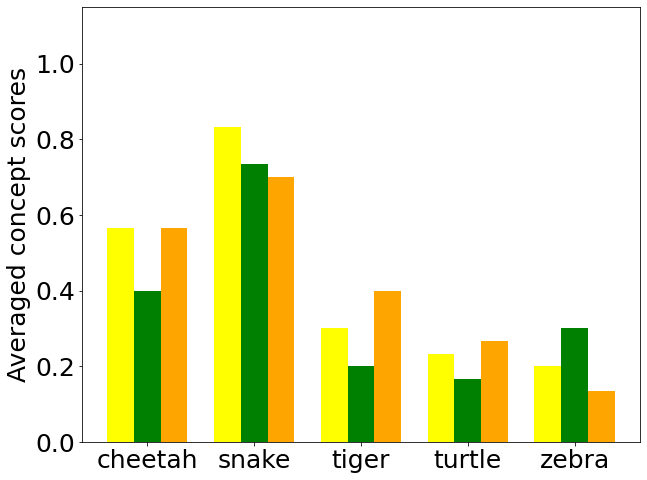}}
    \subfloat[IBD]{\includegraphics[width=\wPlotsProtos\textwidth]{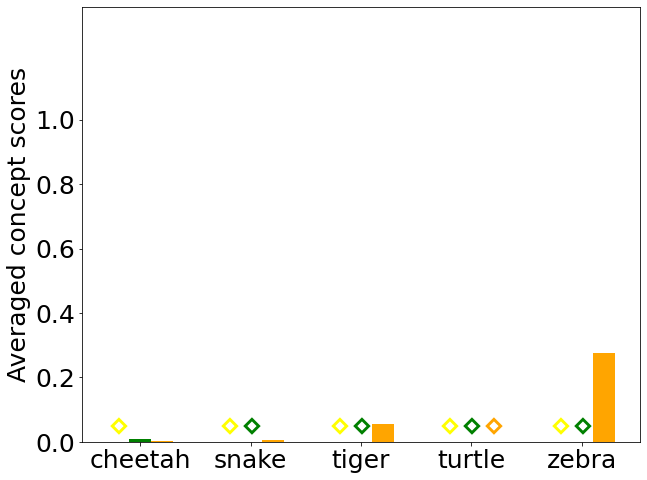}}
    \hfill
    \subfloat[CoProNN]{\includegraphics[width=\wPlotsProtos\textwidth]{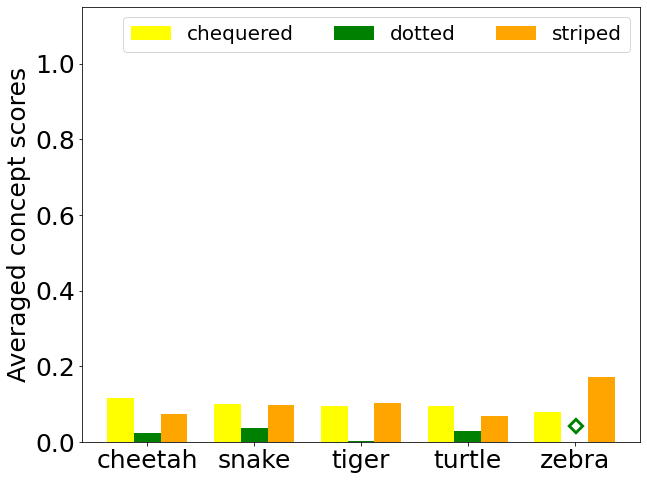}}
    \subfloat[TCAV]{\includegraphics[width=\wPlotsProtos\textwidth]{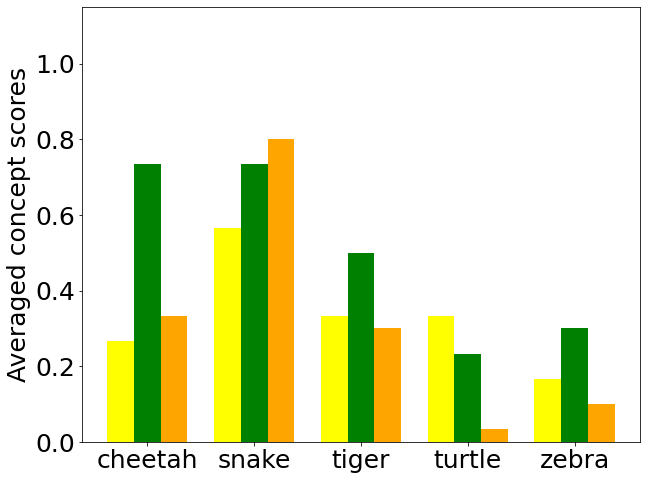}}
    \subfloat[IBD]{\includegraphics[width=\wPlotsProtos\textwidth]{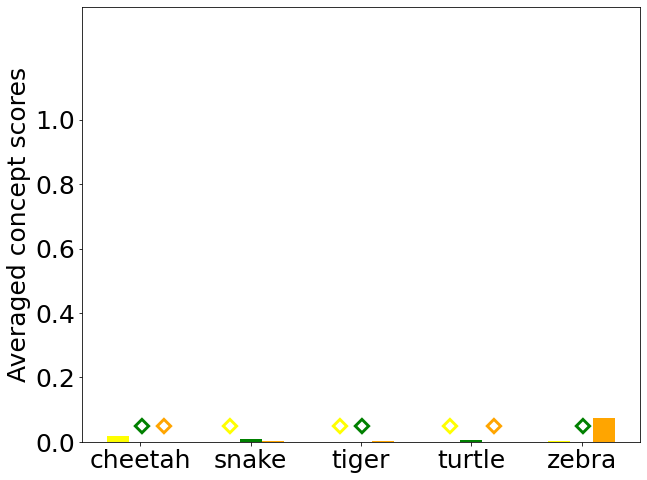}}
    \caption{Low-level task-unspecific concept images do not work well with any XAI method. Top row: methods were applied w.r.t. \textit{broden} concept images. Bottom row: methods were applied w.r.t. SD-computed concept images. A diamond $\diamond$ marks a zero-value. All plots show the average concept relevance scores per class.}
    \label{fig:imagenet_con}
\end{figure*}

For the ImageNet animal classes described in Section \ref{sec:datasets} we also investigated the efficiency of pre-annotated generic and task unspecific concept images from the \textit{broden} dataset when computing explanations with our method, TCAV and IBD. Deeper scrutiny of the \textit{broden} concepts revealed that they were sampled from very different distributions. While the concept set for 'striped' consisted of image patches from the animal world (tiger stripes, fish stripes and so on), the sets for 'dotted' and 'chequered' were either images of clothing, materials or generated with a drawing software. This makes it difficult for the concepts to faithfully represent task relevant features in a given data set or task. To illustrate this effect, we also generated concepts similar to the generic \textit{broden} concepts using their label names as prompts. Yet, these concepts also proved inappropriate for capturing the semantic essence of the animal classes. The results are summarised in Fig. \ref{fig:imagenet_con} and Table \ref{table:concepts} from the Supplement. 
They confirm previous findings  \cite{ramaswamy2023overlooked} highlighting that concept-based explanation methods are sensitive to the concept dataset and that different probe datasets can yield very different explanations. 

\subsection{Quantitative User Study}\label{sec:user_result}

\begin{figure*}[h!]
    \begin{center}
       \includegraphics[width=1.0\textwidth]{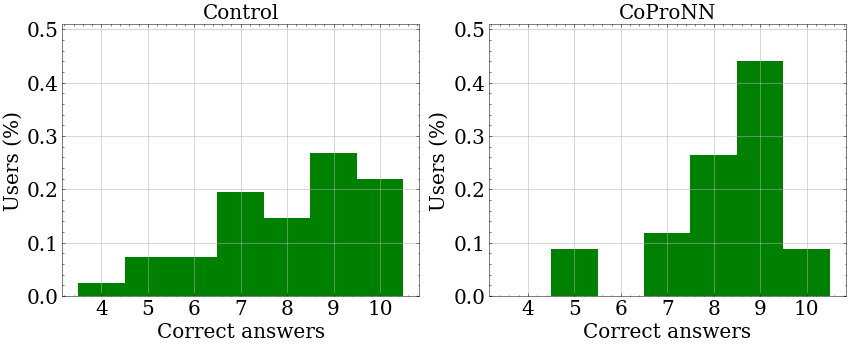}
    \end{center}
    \caption{Explanations improve human performance. There is a trend towards higher classification accuracy when subjects are given explanations for the AI's prediction as opposed to only being shown the unexplained model prediction.}
\label{fig:task2_perform}
\end{figure*}

When analyzing the performance in both groups, we notice higher accuracy in the CoProNN group than in the Control Group (see Fig. \ref{fig:task2_perform}). For orientation, 79\% of the CoProNN participants gave at least 8 correct answers, as opposed to only 63\% in the Control Group. On average, CoProNN users had an accuracy of $8.24\pm1.28$ correct clicks (median at 9 correct clicks), while Control users had an accuracy of $8.05\pm1.64$ (median at 8 correct clicks). 

Recall that two of the ten samples shown to the participants were misclassified by the model. Therefore, users in the Control Group would receive a misleading species suggestion, while CoProNN users would also be shown a misleading explanation for that wrong prediction. We observe an average of 0.59 $\pm$ 0.59 clicks on the wrong model suggestion in the Control Group (median at 1 click) and 0.47 $\pm$ 0.61 clicks in the CoProNN Group (median at 0). Users in the Control Group agreed on 29.27\% of the images with wrong model predictions and users from the CoProNN Group with only 23.53\%. This demonstrates that CoProNN explanations enable users to identify wrong model predictions.

\subsection{Results Qualitative User Study}
\begin{figure*}[h!]
    \begin{center}
       \includegraphics[width=1.0\linewidth]{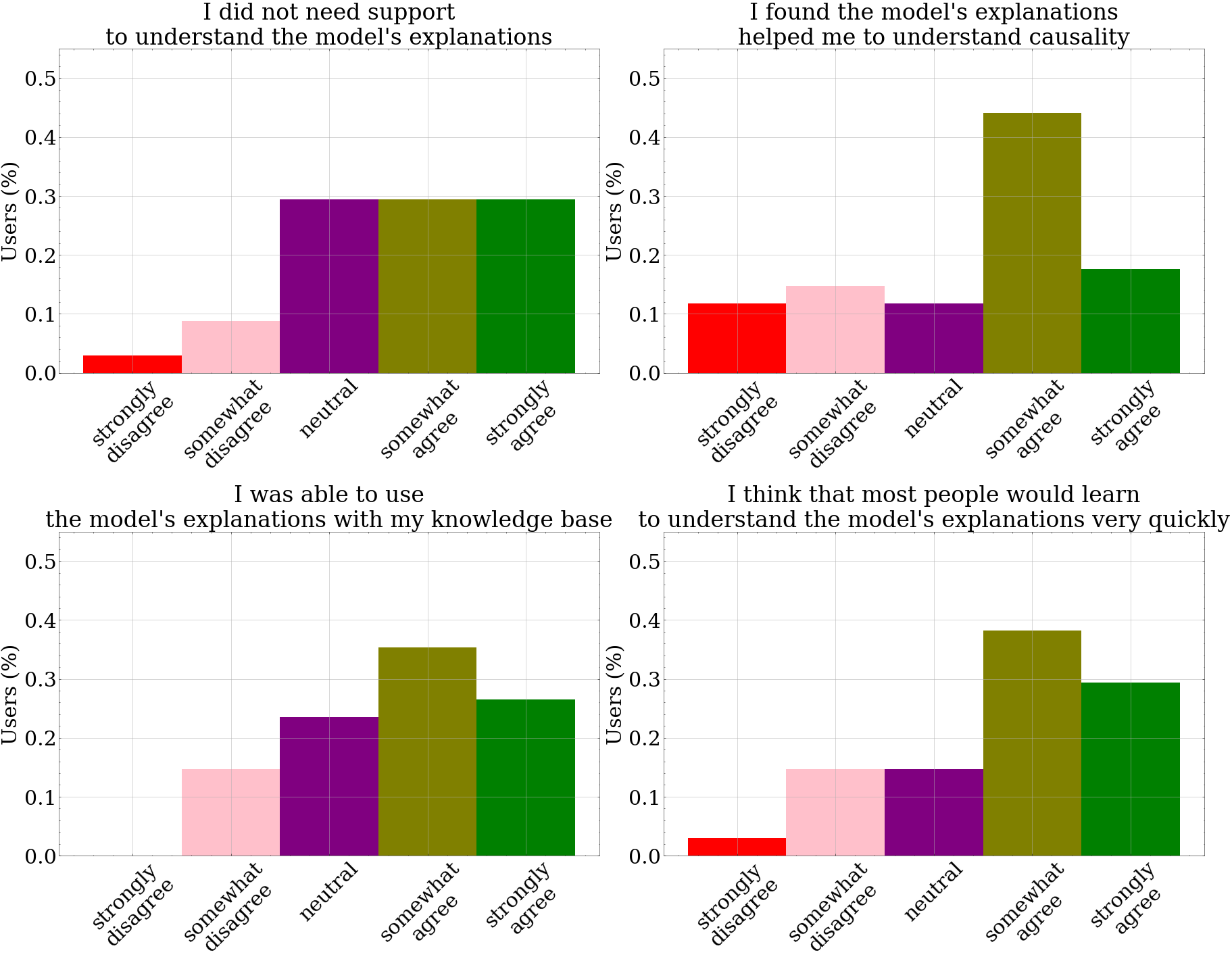}
    \end{center}
    \caption{Results of a survey for qualitative evaluation of CoProNN show that subjects generally found the explanations delivered by our model helpful and easy to understand.}
\label{fig:survey}
\end{figure*}

The results of the survey with 34 participants shown in Figure~\ref{fig:survey} demonstrate that subjects generally found the explanations delivered by our model helpful and easy to understand. 58.82\% claimed that they understood them without further instructions. 61.76\% could grasp causal relationships through the explanations i.e. \textit{"if concept X and Y, but not Z, then species A"}; roughly a quarter of the respondents found that difficult. 61.76\% were able to use the explanations with their prior knowledge (about basic taxonomies in nature) and about two thirds agree that the explanations are accessible to learn by most non-experts (in AI or entomology).

\section{Discussion}

We discuss here the results presented in the previous section. We first take a look at why CoProNN prototypes are more task specific than low-level concepts. Afterwards, we analyze the results gathered from our user study and outline several recommendations for users interested in adapting our method to their own use case. Finally, we point out some of CoProNN's limitations. 

\subsection{Improved Task Specificity of CoProNN Concepts}
Comparing the concepts used in existing methods with the proposed CoProNN approach, we find that concepts created with text-to-image generative Deep Learning methods are simpler to adapt to novel tasks. Concepts such as 'striped' or 'dotted' used in existing methods are often not task-specific and, as such, not representative for the key traits of the considered classes. We demonstrated in our experiments that custom-tailored concepts associated with relevant prototypes are better suited for generating task-specific explanations (see Fig. \ref{fig:specific}). Concepts generated in the CoProNN approach are easily adapted via prompting with specific context to novel scenarios, making them well-suited for versatile use.

\def\wPlotsSpecif{0.25}
\begin{figure*}[h!]
    \centering
    \subfloat[Broden]{\includegraphics[width=\wPlotsSpecif\textwidth]{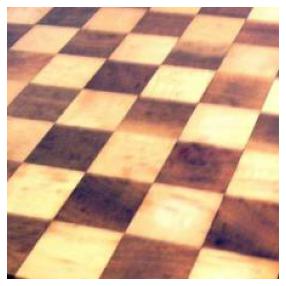}}\hfill
    \subfloat[SD concept]{\includegraphics[width=\wPlotsSpecif\textwidth]{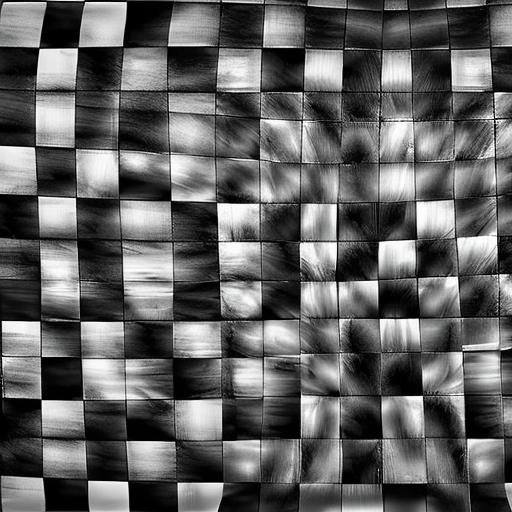}}\hfill
    \subfloat[SD prototype]{\includegraphics[width=\wPlotsSpecif\textwidth]{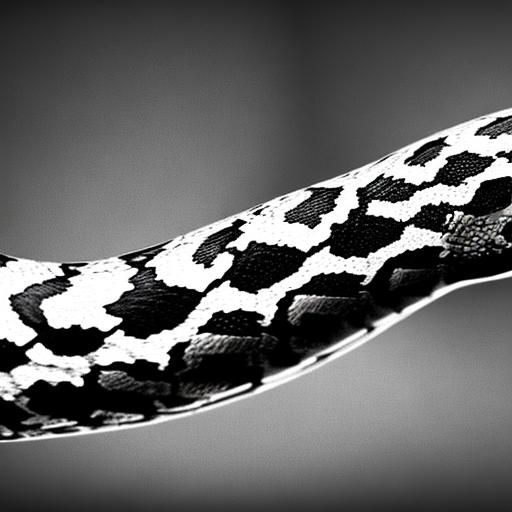}}\hfill
    \hfill
    \caption{High-level concept-based prototypes allow for higher task specificity. Image (a) is taken from the \textit{broden} dataset under the concept set 'chequered'. Image (b) is generated with SD by using a simple prompt condensed in the name of the concept. Image (c) is generated with SD by inputting a more specific prompt, namely 'elongated chequered reptile'. The concept sets from which these examples come were supposed to explain the ImageNet class 'snake'.}
    \label{fig:specific}
\end{figure*}

\subsection{Interpretation User Study Results}
\begin{enumerate}
    \item[1)] \textbf{CoProNN explanations facilitate human-AI collaboration.} The results presented in Section \ref{sec:user_result} and Figure \ref{fig:task2_perform} suggest that CoProNN explanations improved human performance on the classification task.
    \item[2)] \textbf{CoProNN explanations help spotting wrong predictions.} One possible reason for the increased annotation performance in the CoProNN Group would be that the explanations helped users spot wrong predictions more easily. Based on the visual inconsistency of the explanations with the visual appearance of the test sample, CoProNN users manage to identify false predictions more often. It appears, therefore, that explanations enable subjects to better calibrate their trust in the assistive AI, increasing users' critical assessment of the validity of the model's proposals. This contributes to fewer wrong answers in the CoProNN Group.
    \item[3)] \textbf{CoProNN explanations are easy to understand and use.} As the survey answers in Figure \ref{fig:survey} show, respondents generally found the explanations delivered by our CoProNN intuitive and helpful for solving the tasks.
\end{enumerate}

\subsection{Applying CoProNN to Your Own Use Case}
In accordance to the recommendations in \cite{ramaswamy2023overlooked}, we advise practitioners interested in applying our method to concentrate on a relatively small set of intuitive and easily learnable concepts, in order to avoid overwhelming the end-user with the generated explanations. The form in which the explanation is presented to the user is also of paramount importance. We refer the reader to the user study in \cite{Kim_2023}, where the authors \textit{"...found that participants desire practically useful information that can improve their collaboration with the AI, more so than technical system details."}

\subsection{Limitations and Extensions}
Nonetheless, CoProNN has several limitations. First and foremost, our method requires visual concepts that are known and easy to pinpoint. These are key to a successful usage of CoProNN. If the curated concept set is insufficient, our method will have difficulties in producing accurate or useful explanations of model predictions. The concept set is insufficient if not all classes can be uniquely recognized based solely on those concepts. It would mean that the underlying hierarchy does not end in one class per leaf. In that case, the explainer would need to communicate that ambiguity in a proper way to the user. 

Secondly, CoProNN currently works best for tree-structured class domains with conjunctive ("and") characteristics. Especially in a highly complex use-case such as recognizing rare or very similar insect species, disjunctive ("or") relationships are also sometimes necessary for accurate classification. For instance, very often the males and females of the same species have very different characteristics. Accurately assigning them to the correct class would require modelling more complex relationships between these characteristics. Furthermore, not all concepts are necessarily equally important for discriminating a class. For instance, one can say that the "fuzzy orange" abdomen tip of the \textit{Bombus pratorum} is a weaker concept than the dominating trait "yellow black stripes". There are several approaches for addressing this issue such as weighting concepts differently for each class, reformulating CoProNN into a multi-label explainer, precision-recall thresholds and so on.

While the DNN does not require any retraining for CoProNN to work, the explanation method itself does have several hyperparameters that may be used for fine-tuning and adapting the method to other usecases: number of neighbors $k$, threshold $t$, choice of random images, number of random partitions $\alpha$, size of random partitions $\beta$, number of prototypes per concept $n_j$, text prompts for generating the prototypes.  

Finally, CoProNN explanations are better and more robust if the concept sets are well separated/clustered in the search space. One might consider applying a suitable dimensionality reduction technique to the latent space to increase the efficiency of the kNN search.

\section{Conclusion}

In this paper we proposed a novel XAI approach capable of generating intuitive explanations via task-specific concepts. Our method performs very well in comparison to similar concept-based methods, facilitates human-AI collaboration and offers a high degree of flexibility when applied in other scenarios, too. All in all, we are confident that CoProNN offers a viable extension of the current state-of-the-art concept-based XAI approaches and has the potential to be further improved, in order to reliably handle more diverse tasks.

\section*{Acknowledgments}
We thank Christian Schmid-Egger for valuable expert advice on entomological classification of wild bee species.

\bibliographystyle{splncs04}
\bibliography{references}

%%%%%%%%%%%%%%%%%%
%\newpage
\section*{Appendix}

\renewcommand{\algorithmicrequire}{\textbf{Input:}}
\renewcommand{\algorithmicensure}{\textbf{Output:}}
\begin{algorithm}[ht]
    \caption{CoProNN (\textbf{Co}ncept-based \textbf{Pro}totypical \textbf{N}earest \textbf{N}eighbors)}
    \label{alg:knn_concepts}
    \begin{algorithmic}[1]
        \Require \\
            feature extractor $f: \mathbb{R}^{h \times w \times 3} \rightarrow \mathbb{R}^D$, mapping an input image of height $h$ and width $w$ to a feature vector of length $D$\\ 
            m prototypes (explanations), where each prototype $j$ is represented by a set  $C_j = \{c_j^{(1)}, ..., c_j^{(n_j)}\}$ of $n_j$ prototypical images $c$ of dimension $h \times w \times 3$ \\
            $n_{m+1}$ random images $C_{m+1} = \{c_{m+1}^{(1)}, ..., c_{m+1}^{(n_{m+1})}\}$ to extend the search space for kNN \\
            $s$ sample images $\{x_1, ..., x_s\}$ for which to predict the explanations (relevant prototypes) \\
            number of neighbors $k \in \mathbb{N}$ to be considered in kNN \\
            threshold $t \in (0, 1]$ to select the concept candidates for formulating the final explanation \\
            number of partitions $\alpha \in \mathbb{N}$ to sample the random dataset \\
            size of each random partition $\beta < n_{m+1}$

         \Ensure $R = \{R_1, ..., R_s\}$, where $R_i \subset \{1, ..., m\}$ is the set of relevant prototype indices for the test sample $x_i$ 
        
        \Statex
        \Statex \textbf{Extract features of all prototype and random images}
        \For{j in 1 ... m+1} 
            \State $\Omega_j \gets f(C_j) = \{ f(c_j^{(1)}), \dots, f(c_j^{(n_j)}) \}$
        \EndFor
        \Statex

        \Statex \textbf{Run kNN for every random partition}
        \State Initialize zero-matrix $P \in \mathbb{R}^{s \times {(m+1)}}$
        \For{a in 1 ... $\alpha$}
            \State Sample random partition $\Omega_{m+1}^{a}$ of size $\beta$
            \State Fit kNN on $\Omega_1 \cup .... \cup \Omega_m \cup \Omega_{m+1}^{a}$
            \For{i in 1 ... s}
                \State $N(f(x_i)) \gets$ kNN.predict( $f(x_i)$ ) \Comment{set of $k$ nearest proto./random vectors}
                \For{j in 1 ... m}
                    \State $P_{i, j} \gets \cfrac{|N(f(x_i)) \cap \Omega_j|}{k}$ \Comment{ratio of neighbors belonging to $j$-th proto. set}
                \EndFor
            \EndFor
        \EndFor
        \State $P \gets \cfrac{1}{\alpha} P$ \Comment{average scores over all partitions}
        \Statex

        \Statex \textbf{Select relevant prototypes for final explanation}
        \State Initialize empty set $R$
        \For{i in 1 ... s}
            \State Initialize empty set $R_i$
            \For{j in 1 ... m}           
                \If{$P_{ij} \geq t$}
                    \State Append $j$ to $R_{i}$
                \EndIf
            \EndFor
            \State Append $R_{i}$ to $R$
        \EndFor

        \State \textbf{return} $R$
    \end{algorithmic}
\end{algorithm}

%\newpage
\subsection*{Training the Classifier and Optimizing the Explainer}
\label{ap:explainer}
For all three XAI methods (CoProNN, TCAV, IBD), the same random set of 1000 images from the \textit{broden} dataset was used. The random images are not generated with SD, in order to ensure a fair comparison between all three methods, instead of overfitting/fine-tuning on our own method. Indeed, there is plenty of room for improvement when choosing a tailored random dataset for CoProNN. One may argue that generating the random images with SD, too, guarantees a more uniform distribution of all the images used for searching for prototypical neighbors. 

Inspired by the TCAV approach, we implemented the partitioning of this random dataset ($\alpha = 100$ samplings of size $\beta = 30$ without replacement for CoProNN; $\alpha = 30$ samplings of size $\beta = 500$ without replacement for TCAV and IBD) and run the methods multiple times w.r.t. to these partitions, averaging the relevance scores in the end. For CoProNN, the parameter $\beta$ matched the size of every prototype set.

In the TCAV, when computing the Concept Activation Vectors for the 'random concept', we split the random partitions in two further sub-partitions of size 250 (one for the positive class, one for the negative one). 

In IBD, we left out the concept localization part via feature attribution maps and only used the scores resulted from projecting the feature vector $a$ of a test sample on the concept components $q_{c_j}$ that approximate their class weight vectors $\omega_k$ of class $k$. For the sake of comparability to the other two methods, we normalized the scores for every test sample by the corresponding class logits $w_k^T a$ predicted by the classifier (refer to Equations 7-9\footnote{Note that in this paper we used $m$ as the number of considered concepts, which corresponds to $n$ in the cited paper.} in \cite{Zhou_2018_ECCV}):

$$
1 = \underbrace{\cfrac{s_1 q_{c_1}^T a }{w_k^T a}}_\textrm{score of concept $c_1$} + \dots + \underbrace{\cfrac{s_m q_{c_m}^T a }{w_k^T a}}_\textrm{score of concept $c_m$} + \quad r^T a 
$$

We mention here that 
%in the early development stages of 
CoProNN may also be implemented as a multi-label predictor version of the algorithm (each prototype set tested separately against the random set in the kNN, more similar to the implementation of the traditional TCAV), as well as a version without any random set at all (only the concept sets form the search space for the kNN, similar to the 'relative TCAV' implementation). 

Moreover, we tried generating prototypical images in the spirit of the \textit{broden} concepts categorisation (in our case, \textit{color} = \{brown, yellow, orange\} and \textit{texture} = \{smooth, fuzzy\} for the wild bees) and fit two kNNs for each concept category. It turned out that this alternative version would not perform as well as the CoProNN version described in this paper on the aforementioned metrics. More details on these experimental trials are available in our code repository.

\subsubsection*{ImageNet Animals:}
For the ImageNet classes, a frozen ResNet50 \cite{resnet} initialized with ImageNet weights from the TensorFlow library was used. For every class, 150 images were downloaded, out of which 100 were randomly selected as a validation set for optimizing the $k$ parameter in the kNN-explainer. The best $k$-value that would maximize the CS over all classes was 8 when the concept set consisted of SD-generated prototypes and 36 when it consisted of low-level concept images (from \textit{broden} or SD-generated). The remaining 50 images per class made up the test set on which the XAI experiments were conducted.

\subsubsection*{iNat Wild Bees:}
From all the 30k scraped images, an extra subset of 30 images per species was set apart and used for manual annotation\footnote{We note that at the time of building this annotated dataset, powerful off-the-shelf segmenters such as SAM \cite{kirillov2023segment} were not yet released.} via segmentation masks in \textit{Label Studio} \cite{lstudio}. Our annotated dataset is freely available online at \href{https://zenodo.org/record/6642157}{https://zenodo.org/record/6642157}. The masks were then used to train a MaskRCNN \cite{maskrcnn}, with which the remaining scraped images were automatically segmented\footnote{The pipeline for training the segmenter was implemented via a forked repository: \href{https://github.com/lucasjinreal/yolov7_d2}{https://github.com/lucasjinreal/yolov7\_d2}.}. The MaskRCNN-segmented images belonging only to the five species of interest were then cropped down to their minimal bounding box, split into a training and validation set (2/3 - 1/3) and fed into a ResNet50 initialized with a backbone \cite{Horn} pretrained on the 2021 iNat Challenge Dataset \cite{iNat}. The images of the five species manually annotated in Label Studio were set aside as test set for the classifier. We report a \textbf{0.85 top-1} test accuracy. Similar to \cite{Horn}, we conducted first an experiment to test which input format - raw image, segmented or segmented + cropped to bounding box - leads to the highest model performance. The third format proved to deliver the best results.

As far as the kNN-explainer is concerned: we searched for an optimal $k$ on the combined training and validation set. The best value for $k$ w.r.t. CS over all classes was found to be 18.

\begin{figure*}[!ht]
    \begin{center}
       \includegraphics[width=1.0\textwidth]{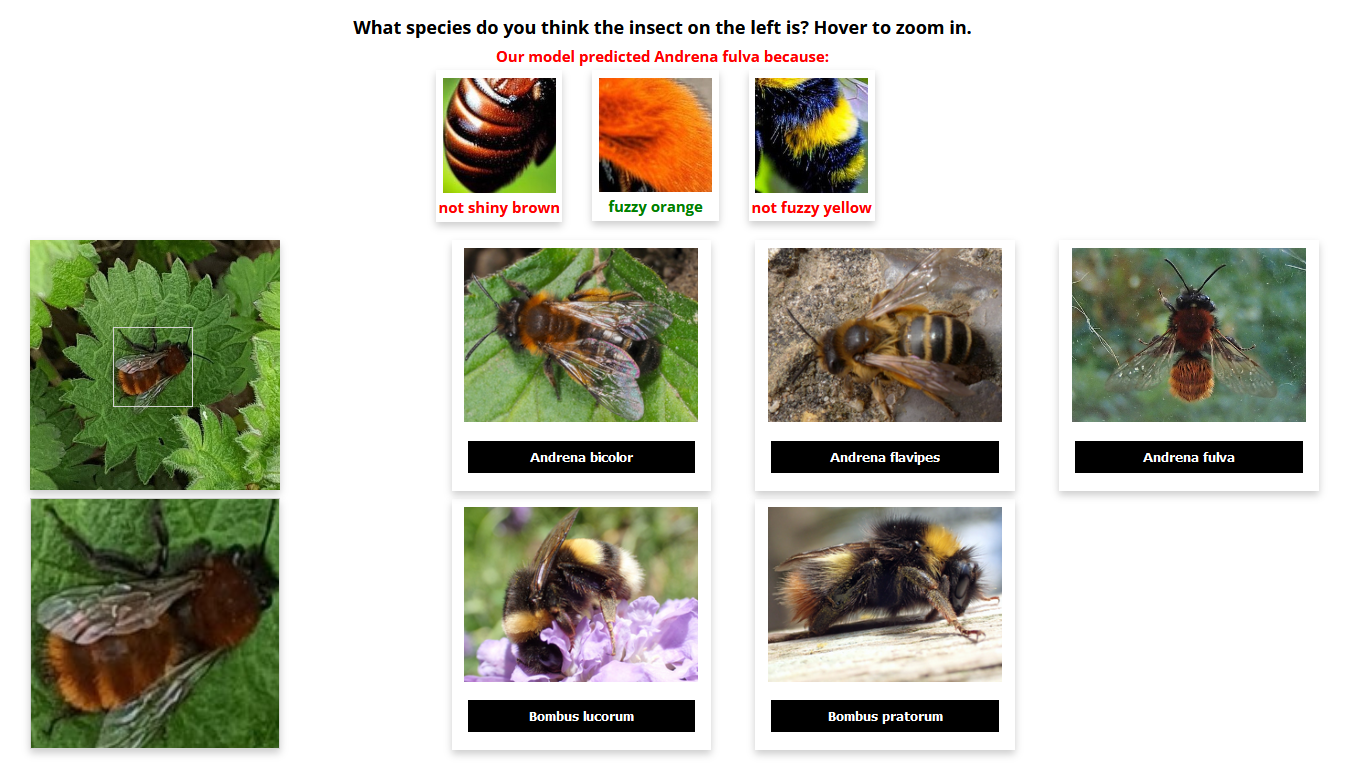}
    \end{center}
    \caption{Example of one task an XAI user may receive. To answer the question, the subjects need to click on one of the five species on the right. Beneath the test sample on the left there is zoom-in pane to magnify certain portions of the image.}
\label{fig:cognition}
\end{figure*}

\begin{figure*}[ht!]
    \centering
    \subfloat[]{\includegraphics[width=0.45\textwidth]{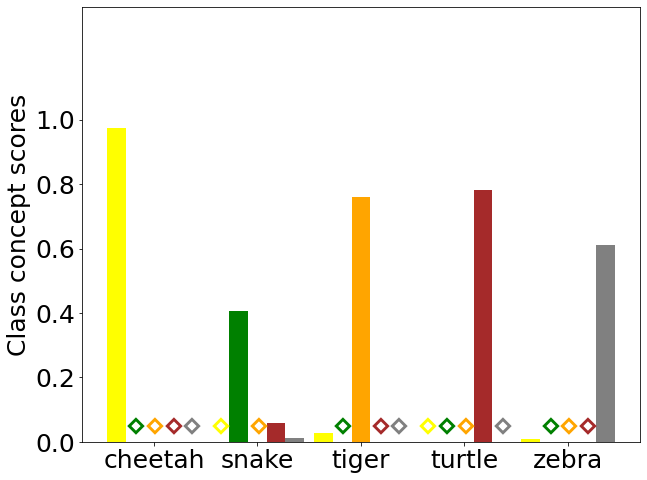}}
    \hfill
    \subfloat[]{\includegraphics[width=0.45\textwidth]{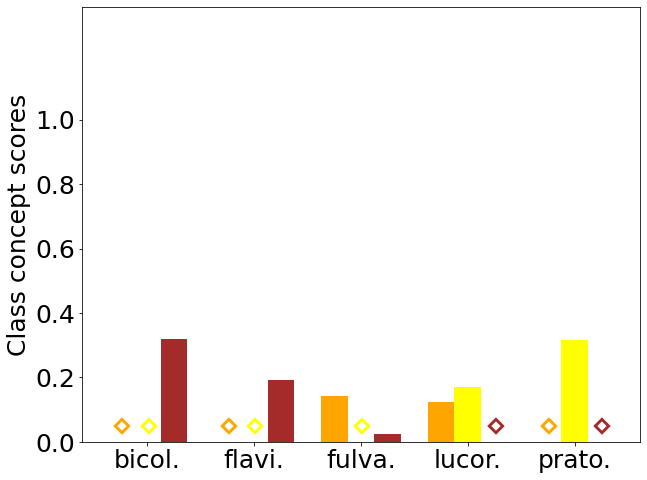}}
    \caption{IBD class scores for the ImageNet animal classes (a) and the iNat wild bees with concept prototypes (b). A diamond $\diamond$ marks a zero-value.}
    \label{fig:ibd_cls_scores}
\end{figure*}

\begin{table}[h!]
    \centering
    \begin{tabular}{|m{0.3em } | m{3.5em} || m{8em} | m{8em} | m{8em} |} 
     \hline
      & & \multicolumn{3}{|c|}{\textbf{Cosine Similarity}} \\
     \cline{3-5}
     &\textbf{\textit{Class}} & \textbf{CoProNN} & \textbf{TCAV} & \textbf{IBD} \\
     \hline
     \multirow{5}{*}{\rotatebox{90}{Broden Concepts}}
     &\textit{Cheetah} & 0.1207 $\pm$ 0.1242 & 0.4225 $\pm$ 0.001 & \textbf{0.7316 $\pm$ 0.6209} \\
     \cline{2-5}
     &\textit{Snake} & 0.3274 $\pm$ 0.0866 & \textbf{0.6572 $\pm$ 0.001} & 0.0 $\pm$ 0.0 \\
     \cline{2-5}
     &\textit{Tiger} & \textbf{0.9973 $\pm$ 0.0084} & 0.7448 $\pm$ 0.001 & 0.9535 $\pm$ 0.3014 \\
     \cline{2-5}
     &\textit{Turtle} & 0.2307 $\pm$ 0.1097 & \textbf{0.5367 $\pm$ 0.001} & 0.0 $\pm$ 0.0 \\
     \cline{2-5}
     &\textit{Zebra} & 0.9978 $\pm$ 0.0035 & 0.3906 $\pm$ 0.001 & \textbf{1.0 $\pm$ 0.0} \\
     \hline\hline
     
     \multirow{5}{*}{\rotatebox{90}{SD Concepts}}
     &\textit{Cheetah} & 0.168 $\pm$ 0.164 & \textbf{0.8835 $\pm$ 0.001} & 0.0 $\pm$ 0.0 \\
     \cline{2-5}
     &\textit{Snake} & \textbf{0.664 $\pm$ 0.1403} & 0.5958 $\pm$ 0.001 & 0.0 $\pm$ 0.0 \\
     \cline{2-5}
     &\textit{Tiger} & \textbf{0.718 $\pm$ 0.1231} & 0.6058 $\pm$ 0.001 & 0.1628 $\pm$ 0.9867 \\
     \cline{2-5}
     &\textit{Turtle} & \textbf{0.7696 $\pm$ 0.0941} & 0.6046 $\pm$ 0.001 & 0.0 $\pm$ 0.0 \\
     \cline{2-5}
     &\textit{Zebra} & 0.8915 $\pm$ 0.0561 & 0.4634 $\pm$ 0.001 & \textbf{0.9999 $\pm$ 0.0001} \\
     \hline
    \end{tabular}
    \caption{Low-level task-unspecific concept images do not work well with any XAI method, neither on pre-annotated \textit{broden} concepts, nor on similar concepts generated with SD. CoProNN appears to efficiently explain the striped animals (tiger and zebra) on the \textit{broden} concepts, but plot (a) in Fig. \ref{fig:imagenet_con} reveals that the 'striped' concept is always predicted as dominant regardless of the class. Also, on the SD concepts, TCAV only explains the cheetah class well, fact highlighted in plot (e) of Fig. \ref{fig:imagenet_con} where only the cheetah has a clear peak for the correct 'dotted' concept.}
    \label{table:concepts}
\end{table}

\end{document}